\documentclass{article}

\usepackage[preprint]{neurips_2026}

\usepackage[utf8]{inputenc}
\usepackage[T1]{fontenc}
\usepackage{hyperref}
\usepackage{url}
\usepackage{booktabs}
\usepackage{amsmath}
\usepackage{amsfonts}
\usepackage{microtype}
\usepackage{xcolor}
\usepackage{graphicx}
\usepackage{placeins}

\title{Will the Judge Flip? Predicting Position-Sensitive LLM Judgments from Residual Stream Activations}

\author{
  Hashmath Shaik \qquad Gnaneswar Villuri \qquad Alex Doboli \\
  Department of Electrical and Computer Engineering \\
  Stony Brook University \\
  Stony Brook, NY 11794, USA \\
  {\small\texttt{\{hashmath.shaik,gnaneswar.villuri,alex.doboli\}@stonybrook.edu}}
}

\hypersetup{
  pdftitle={Will the Judge Flip? Predicting Position-Sensitive LLM Judgments from Residual Stream Activations},
  pdfauthor={Hashmath Shaik, Gnaneswar Villuri, Alex Doboli}
}

\begin{document}

\raggedbottom

\maketitle

\begin{abstract}
The order in which candidate responses are presented can change an LLM judge's verdict.
Detecting such a \emph{position flip} ordinarily requires judging each pair in both orders, which doubles the number of judgments.
We investigate whether residual stream activations recorded immediately before the initial verdict can predict a flip.
We use nested grouped cross-validation to evaluate regularized linear probes on 534 JudgeBench pairs for three Qwen3 judges and Llama-3.1-8B.
The linear probes achieve AUROCs of .621--.850 and outperform a combined baseline that uses verbalized confidence, verdict-label logits, response lengths, and the judge's initial choice by .062--.113 AUROC.
Linear probes trained on JudgeBench and then frozen achieve AUROCs of .685--.853 on 1,802 MT-Bench comparisons without MT-Bench fitting or recalibration.
These results show that pre-verdict activations support prediction of susceptibility to candidate order and outperform the non-activation predictors evaluated here.
\end{abstract}

\section{Introduction}

LLM-based judges compare model responses, but their verdicts can depend on candidate order \citep{zheng2023judging,wang2024fair}.
A \emph{position flip} occurs when exchanging the candidates changes the judge's preferred response identity.
We denote the order with A first as $AB$ and the swapped order with B first as $BA$.
Such flips are associated with less reliable judgments: for our four judges, the $AB$ verdict agrees with the JudgeBench reference on 60--69\% of non-flipping pairs but only 47--51\% of flipping pairs.\footnote{These are post-hoc accuracies for the $AB$ evaluation. On a flipping pair, the $AB$ and $BA$ verdicts necessarily select opposite response identities, so their accuracies sum to one; this comparison does not imply that exchanging positions reduces accuracy.}

Identifying a position flip conventionally requires evaluating the same pair twice, once in each order.
This procedure detects positional sensitivity only after the $BA$ judgment, which it requires for every pair, including those whose verdict remains unchanged.
We therefore ask whether information available during the $AB$ evaluation can predict which comparisons are susceptible to a flip.

Our predictor is deliberately simple: an L2-regularized linear probe trained on a residual stream activation recorded immediately before the $AB$ verdict.
The linear probe receives no information from the $BA$ evaluation, which is used only to construct the position-flip label.

We address three research questions:
\textbf{RQ1:} Can a residual stream activation from the $AB$ evaluation predict whether the judgment will flip?
\textbf{RQ2:} Does this representation predict flips better than non-activation features from the $AB$ evaluation, including the judge's $AB$ choice?
\textbf{RQ3:} Does a linear probe trained on JudgeBench transfer to MT-Bench without adaptation?

Across all four judges, the linear probes predict held-out position flips and outperform the combined non-activation baseline.
The signal also transfers to MT-Bench without retraining or recalibration.
These results establish linear decodability, but they do not show that the model uses the decoded information causally or explicitly estimates positional susceptibility.

\paragraph{Related work.}

Prior work measures positional bias by comparing judgments across candidate orders.
MT-Bench documents this effect \citep{zheng2023judging}, and subsequent work shows that candidate order can alter model rankings \citep{wang2024fair}.
Related studies examine the consistency, fairness, and stability of reordered judgments \citep{shi2025judging}, while CalibraEval adjusts prediction distributions to reduce disagreement after candidate swaps \citep{li2025calibraeval}.
These studies characterize or mitigate positional sensitivity.
We instead seek to predict an individual flip from the $AB$ evaluation alone.

Internal representations have also been used to estimate judge reliability.
Hidden-state probes can capture correctness uncertainty \citep{radharapu2025calibrating}, and cross-layer representations can improve agreement between pointwise judge scores and human ratings \citep{lai2025beyond}.
Recent work also recovers human-preference signals from judge activations and examines internal pathways carrying position and length bias \citep{mukherjee2026opening}.
Our target differs from correctness, human-preference recovery, and score alignment: we predict whether the judge's own preferred response identity will change under an order swap.
Probe performance establishes only that a property is decodable, not that the model uses it causally \citep{hewitt2019probes}.
We therefore interpret our results strictly as predictive evidence.

\section{Method and Experimental Setup}

\paragraph{Position-flip target.}
Each instance contains a user request $x$ and two responses, A and B.
We map each verdict to the selected response's original identity, yielding $c_{AB},c_{BA}\in\{A,B\}$; A and B denote response identities rather than presentation slots in both orders.
The position-flip label and aggregate position flip rate (PFR) are
\begin{equation}
y_{\mathrm{pos}}=\mathbf{1}[c_{AB}\ne c_{BA}],
\qquad
\mathrm{PFR}=\frac{1}{N}\sum_{i=1}^{N}y_{\mathrm{pos}}^{(i)}.
\label{eq:target}
\end{equation}
Thus, $y_{\mathrm{pos}}=1$ means that exchanging the candidates changes the selected response.
The label captures susceptibility to candidate order but does not distinguish a consistent preference for the first or second candidate from position effects specific to a particular comparison.

\paragraph{Residual stream activations and linear probe.}
The linear probe uses a residual stream activation recorded immediately before the verdict token.
The fixed prompt requests brief reasoning, an integer confidence score from 0 to 100, and a final single-token verdict.
We use greedy decoding with a 256-token generation allowance and disable Qwen3 thinking mode.
To extract activations from the $AB$ evaluation, we replay the saved token sequence through the same token-by-token KV-cache path used during generation.
We record this activation at four predeclared relative depths: 25\%, 50\%, 75\%, and the penultimate layer.
Let $h_{AB}$ denote the activation at a given layer.
For every example in the shared JudgeBench analysis set, replay reproduces the recorded verdict as the greedy next token.
We train the linear probe as an L2-regularized logistic regression classifier:
\begin{equation}
\hat p(y_{\mathrm{pos}}=1\mid h_{AB})=\sigma(w^\top h_{AB}+b).
\label{eq:probe}
\end{equation}
We optimize the linear probe with deterministic L-BFGS and select both the layer and $\lambda\in\{10^{-4},10^{-3},10^{-2},10^{-1},1,10\}$ by validation Brier score.
The $BA$ evaluation contributes only to the target in Eq.~\ref{eq:target}; it supplies no feature to the linear probe.

\paragraph{Baselines.}
All baselines use only information from the $AB$ evaluation and are evaluated on the same held-out folds as the linear probe.
The confidence baseline uses the judge's verbalized confidence score.
The logit-margin baseline uses $z_A-z_B$ and $|z_A-z_B|$, where $z_A$ and $z_B$ are the next-token verdict logits mapped to responses A and B.
The length-difference baseline uses the signed and absolute differences between the token lengths of responses A and B.
The $AB$-choice baseline indicates whether the $AB$ verdict selects A or B.
The \emph{combined baseline} uses all six features: confidence; signed and absolute logit margin; signed and absolute length difference; and $AB$ choice.
The \emph{baseline without $AB$ choice} omits this feature, and a no-signal predictor uses the training-fold PFR.
Every learned baseline follows the same nested fitting procedure as the linear probe.

\paragraph{Data and judges.}
JudgeBench \citep{tan2025judgebench} is our in-domain benchmark.
Our JudgeBench analysis pool contains 620 pairs.
We use a 32,768-token context budget in both candidate orders.
Of these, 534 pairs from 470 question groups are usable for every judge, which keeps the evaluated examples identical across judges and predictors.
Appendix Table~\ref{tab:retention-control} reports per-judge retention.

MT-Bench \citep{zheng2023judging} provides the out-of-domain transfer data.
Deduplication yields 2,400 comparisons across 80 questions.
The human-judgment split adds no unique comparisons, and we do not use preference labels.
Each question contains 30 comparisons, so we group examples by question.
Of these, 1,802 comparisons from 79 questions are usable for every judge; Appendix Table~\ref{tab:retention-control} reports per-judge retention.
Llama follows the required three-line format less consistently, although 99.4\% of its outputs reach a verdict token.
Our evaluation prompt differs from the official FastChat templates, so the experiment tests transfer across response-pair distributions and does not yield official-protocol PFR estimates.

We evaluate Qwen3-1.7B, Qwen3-4B, and Qwen3-8B in non-thinking mode \citep{yang2025qwen3}, together with Llama-3.1-8B-Instruct \citep{grattafiori2024llama3} as a cross-family replication.
We generate all judgments deterministically in bfloat16 on four 48-GB NVIDIA RTX A6000 GPUs.
Independent reproduction additionally requires the exact checkpoint revisions, prompts, parser rules, and split identifiers.

\paragraph{Evaluation.}
On JudgeBench, we use five outer grouped folds and three inner grouped folds shared across all judges and predictors.
Grouping keeps variants of the same question together, while feature standardization and model selection are performed exclusively within the relevant training folds.
We concatenate predictions from the five held-out outer folds, producing exactly one held-out prediction per comparison.

For MT-Bench transfer, grouped cross-validation over all 534 JudgeBench pairs selects the hyperparameters.
We then refit and freeze one linear probe and its standardization on the complete JudgeBench set.
No fitting, threshold selection, or recalibration uses MT-Bench data.

We report PFR, AUROC, AUPRC, and Brier score \citep{brier1950verification}.
Choice-matched AUROC evaluates rankings only between flipping and non-flipping examples with the same $AB$ choice.
We predeclare that we will not report discrimination metrics when PFR falls below .05 or exceeds .95; no judge--benchmark combination meets either condition.
Percentile 95\% confidence intervals use 2,000 group-bootstrap resamples, with the underlying question as the resampling unit.
Confidence intervals for differences between predictors use paired resamples.

The repeated-$AB$ control measures same-order variation under deterministic inference.
For this control, we select 100 pairs before inference using a seeded hash of the pair identifier.

\paragraph{Code availability.}
The code repository for this study is \url{https://github.com/hashmathshaik/will-the-judge-flip}.

\section{Results}

\subsection{Residual stream activations predict position flips (RQ1)}

Residual stream activations from the $AB$ evaluation predict position flips for every judge tested.
Table~\ref{tab:main} reports one held-out prediction for each of the 534 shared JudgeBench pairs.
All four AUROC confidence intervals exclude .5; point estimates range from .621 for Llama-3.1-8B to .850 for Qwen3-8B.

\begin{table}[htbp]
\caption{Linear-probe performance on 534 shared JudgeBench pairs. Brackets denote 95\% group-bootstrap confidence intervals. Higher AUROC and AUPRC and lower Brier score indicate better performance.}
\label{tab:main}
\centering
\footnotesize
\setlength{\tabcolsep}{3.5pt}
\begin{tabular}{lrrrrr}
\toprule
Judge & Pos. flips/$N$ & PFR & AUROC $\uparrow$ & AUPRC $\uparrow$ & Brier $\downarrow$ \\
\midrule
Qwen3-1.7B   & 269/534 & .504 & .640 [.595,.686] & .615 [.554,.677] & .233 [.224,.241] \\
Qwen3-4B     & 168/534 & .315 & .762 [.719,.804] & .578 [.507,.660] & .178 [.161,.195] \\
Qwen3-8B     & 199/534 & .373 & .850 [.815,.884] & .750 [.686,.812] & .150 [.132,.168] \\
Llama-3.1-8B & 215/534 & .403 & .621 [.569,.669] & .507 [.447,.583] & .234 [.220,.249] \\
\bottomrule
\end{tabular}
\end{table}

\FloatBarrier

\subsection{Residual stream activations outperform the combined baseline (RQ2)}

The linear probe outperforms a combined baseline that includes the judge's $AB$ choice and other non-activation features.
We include $AB$ choice because the pre-verdict activation may encode it, making that choice a simpler potential predictor of flipping.
The baseline also includes verbalized confidence, verdict-label logit margin, and response-length difference.
Linear-probe AUROC exceeds the combined baseline by .062--.113 across the four judges, with every paired confidence interval excluding zero (Appendix Table~\ref{tab:paired-differences}).
The paired intervals do not establish improvements on every metric: the Qwen3-4B AUPRC interval and the Llama-3.1-8B Brier interval include zero.

Choice-matched comparisons confirm that $AB$ choice alone does not explain the probe's discrimination.
Appendix Table~\ref{tab:baselines} reports choice-matched AUROC, which compares flipping and non-flipping examples with the same $AB$ choice.
Under this stricter comparison, the linear probe improves over the combined baseline by .108--.126, and every paired confidence interval excludes zero.
Thus, the probe's discrimination is not explained by $AB$ choice alone and exceeds that of the fitted combined baseline.

\subsection{Transfer to a different benchmark (RQ3)}

Linear probes trained on JudgeBench generalize to MT-Bench without adaptation.
They achieve AUROCs of .685--.853, with every 95\% confidence interval above .5 (Table~\ref{tab:transfer}; Appendix~\ref{app:transfer-ci}).
MT-Bench PFR ranges from .172 to .379, compared with .315 to .504 on JudgeBench.
AUPRC exceeds MT-Bench flip prevalence for every judge, and Brier score is lower than that of a constant predictor fixed to JudgeBench prevalence.
Because the MT-Bench linear probes are refit on all JudgeBench examples, differences between the two table blocks are not paired performance changes.

\begin{table}[htbp]
\caption{JudgeBench results concatenate predictions from five fold-specific linear probes. MT-Bench results use one linear probe refit on all 534 JudgeBench pairs and frozen before transfer.}
\label{tab:transfer}
\centering
\footnotesize
\setlength{\tabcolsep}{3pt}
\begin{tabular}{lrrrrrrrr}
\toprule
& \multicolumn{4}{c}{JudgeBench (out-of-fold)} & \multicolumn{4}{c}{MT-Bench (frozen transfer)} \\
\cmidrule(lr){2-5} \cmidrule(lr){6-9}
Judge & PFR & AUROC $\uparrow$ & AUPRC $\uparrow$ & Brier $\downarrow$ & PFR & AUROC $\uparrow$ & AUPRC $\uparrow$ & Brier $\downarrow$ \\
\midrule
Qwen3-1.7B   & .504 & .640 & .615 & .233 & .379 & .685 & .538 & .234 \\
Qwen3-4B     & .315 & .762 & .578 & .178 & .172 & .735 & .349 & .135 \\
Qwen3-8B     & .373 & .850 & .750 & .150 & .173 & .853 & .529 & .110 \\
Llama-3.1-8B & .403 & .621 & .507 & .234 & .192 & .734 & .401 & .150 \\
\bottomrule
\end{tabular}
\end{table}

\FloatBarrier

\paragraph{Repeated-$AB$ control.}
On a predeclared 100-pair subset, repeating $AB$ changes no verdict, whereas $BA$ changes 19--47 verdicts depending on the judge (Appendix Table~\ref{tab:retention-control}).
This control provides no evidence of same-order verdict variation on the tested subset under the same inference settings.

\section{Limitations}

We test four relatively small open-weight judges, including three Qwen3 models, on two response-pair distributions.
PFR may depend on the judge, data, or prompt; our evaluation prompt also precludes comparison with official MT-Bench estimates.
The shared analysis sets retain 86.1\% of our JudgeBench pool and 75.1\% of the MT-Bench comparisons.
Because inclusion requires valid outputs in both candidate orders for every judge, these results are conditional on format compliance and may not generalize to excluded comparisons.

The combined baseline does not exhaust all measures of difficulty or uncertainty, and the confidence-before-verdict format may weaken verbalized confidence.
Outperforming the fitted linear baseline does not exclude nonlinear predictive relationships among its features.
Other controls or elicitation protocols may therefore change the RQ2 comparison.

Finally, linear decodability does not establish causal use, and residual stream access excludes closed models.
Linear-probe scores predict position-flip risk rather than general reliability; untested settings still require evaluating sampled pairs in both orders.
We do not evaluate a risk-threshold policy, realized judgment savings, or net runtime including activation extraction.

\section{Conclusion}

Pre-verdict residual stream activations predict position flips across the four judges tested.
Linear probes outperform the combined baseline on JudgeBench and transfer to MT-Bench without adaptation.
These results demonstrate predictive signal but not causal use.

\bibliographystyle{plainnat}
\bibliography{references}

\clearpage
\appendix

\section{Data retention and repeated-presentation control}

\begin{table}[htbp]
\caption{Per-judge usable-pair counts and repeated-presentation results. A pair is usable only when both candidate orders satisfy the fixed output format. Repeating $AB$ changes no verdict for any judge on the 100-pair control subset; the final column reports verdict changes after swapping to $BA$.}
\label{tab:retention-control}
\centering
\footnotesize
\setlength{\tabcolsep}{7pt}
\begin{tabular}{lrrr}
\toprule
Judge & JudgeBench usable/$620$ & MT-Bench usable/$2{,}400$ & $BA$ changes/$100$ \\
\midrule
Qwen3-1.7B   & 579 & 2,380 & 47 \\
Qwen3-4B     & 586 & 2,384 & 19 \\
Qwen3-8B     & 590 & 2,396 & 30 \\
Llama-3.1-8B & 547 & 1,831 & 43 \\
\bottomrule
\end{tabular}
\end{table}

\section{Additional baseline results}
\label{app:baselines}

\begin{table}[htbp]
\caption{JudgeBench AUROC on identical held-out pairs and folds. The combined baseline uses all six baseline features; the baseline without $AB$ choice omits that feature. Choice-matched AUROC compares flipping and non-flipping examples with the same $AB$ choice. Brackets denote 95\% paired group-bootstrap intervals for the linear probe's improvement over the combined baseline.}
\label{tab:baselines}
\centering
\footnotesize
\setlength{\tabcolsep}{4pt}
\begin{tabular}{lccccc}
\toprule
Judge & Linear probe & Combined & Without $AB$ choice & Logit margin & Difference \\
\midrule
\multicolumn{6}{l}{\emph{Pooled AUROC $\uparrow$}} \\
Qwen3-1.7B   & \textbf{.640} & .578 & .565 & .577 & +.062 [.018,.107] \\
Qwen3-4B     & \textbf{.762} & .668 & .652 & .657 & +.094 [.054,.135] \\
Qwen3-8B     & \textbf{.850} & .776 & .765 & .743 & +.074 [.041,.109] \\
Llama-3.1-8B & \textbf{.621} & .508 & .469 & .518 & +.113 [.055,.171] \\
\midrule
\multicolumn{6}{l}{\emph{Choice-matched AUROC $\uparrow$}} \\
Qwen3-1.7B   & \textbf{.635} & .527 & .524 & .532 & +.108 [.046,.169] \\
Qwen3-4B     & \textbf{.744} & .620 & .619 & .618 & +.124 [.077,.174] \\
Qwen3-8B     & \textbf{.785} & .660 & .642 & .592 & +.126 [.065,.184] \\
Llama-3.1-8B & \textbf{.614} & .490 & .450 & .509 & +.123 [.064,.182] \\
\bottomrule
\end{tabular}
\end{table}

\begin{table}[htbp]
\caption{Paired differences between the linear probe and combined baseline on JudgeBench. Brackets denote 95\% paired group-bootstrap intervals. Positive AUROC/AUPRC differences and negative Brier differences favor the linear probe.}
\label{tab:paired-differences}
\centering
\footnotesize
\setlength{\tabcolsep}{4pt}
\begin{tabular}{lccc}
\toprule
Judge & AUROC diff. $\uparrow$ & AUPRC diff. $\uparrow$ & Brier diff. $\downarrow$ \\
\midrule
Qwen3-1.7B   & +.062 [.018,.107] & +.079 [.022,.133] & -.008 [-.015,-.001] \\
Qwen3-4B     & +.094 [.054,.135] & +.053 [-.004,.110] & -.018 [-.030,-.007] \\
Qwen3-8B     & +.074 [.041,.109] & +.095 [.030,.155] & -.034 [-.048,-.020] \\
Llama-3.1-8B & +.113 [.055,.171] & +.097 [.038,.161] & -.012 [-.024,.001] \\
\bottomrule
\end{tabular}
\end{table}

\clearpage
\section{Transfer confidence intervals}
\label{app:transfer-ci}

\begin{table}[htbp]
\caption{Performance on 1,802 shared MT-Bench comparisons for linear probes refit on JudgeBench and frozen before transfer. Brackets denote 95\% question-group-bootstrap intervals.}
\centering
\footnotesize
\setlength{\tabcolsep}{4pt}
\begin{tabular}{lccc}
\toprule
Judge & AUROC $\uparrow$ & AUPRC $\uparrow$ & Brier $\downarrow$ \\
\midrule
Qwen3-1.7B   & .685 [.653,.714] & .538 [.495,.584] & .234 [.228,.240] \\
Qwen3-4B     & .735 [.691,.778] & .349 [.301,.410] & .135 [.121,.152] \\
Qwen3-8B     & .853 [.823,.883] & .529 [.461,.605] & .110 [.097,.124] \\
Llama-3.1-8B & .734 [.689,.772] & .401 [.334,.472] & .150 [.139,.162] \\
\bottomrule
\end{tabular}
\end{table}

\end{document}